\documentclass[nohyperref]{article}

\usepackage{amsmath,amsfonts,bm,amssymb}

\def\eqref#1{equation~\ref{#1}}

\def\1{\bm{1}}

\DeclareMathAlphabet{\mathsfit}{\encodingdefault}{\sfdefault}{m}{sl}
\SetMathAlphabet{\mathsfit}{bold}{\encodingdefault}{\sfdefault}{bx}{n}

\newcommand{\R}{\mathbb{R}}

\DeclareMathOperator*{\argmin}{arg\,min}

\usepackage{latexsym,eucal,bbm,color}
\usepackage{url}
\usepackage{comment}
\usepackage{float}
\usepackage{tcolorbox}
\usepackage{booktabs}
\usepackage{blindtext}
\usepackage{hyperref}
\hypersetup{
    colorlinks,
    linkcolor={red!50!black},
    citecolor={blue!50!black},
    urlcolor={blue!80!black}
}
\usepackage{multirow}
\usepackage{booktabs}
\usepackage{array}
\usepackage{soul}
 
\usepackage[normalem]{ulem}
\usepackage{stackengine}
\usepackage{parskip}
\usepackage{bm}
\usepackage{balance}
\usepackage{comment}
\usepackage{soul}
\usepackage{pgffor}

\usepackage{amsthm}
\usepackage[ruled,vlined]{algorithm2e}

\newtheorem{thm}{Theorem}

\usepackage{caption}

\def \bx{\boldsymbol{x}}

\def \by{\boldsymbol{y}}
\def \bz{\boldsymbol{z}}
\def \bm{\boldsymbol{m}}

\def \bp{\boldsymbol{p}}

\def \bb{\boldsymbol{b}}

\def \bm{\boldsymbol{m}}

\def \bA{\boldsymbol{A}}

\def \bC{\boldsymbol{C}}
\def \bZ{\boldsymbol{Z}}

\def\R{{\mathbb R}}

\def\Indic{\mathbbm{1}}

\newcommand{\Aff}{\text{Aff}}

\usepackage{microtype}
\usepackage{graphicx}
\usepackage{subfigure}
\usepackage{booktabs} 

\usepackage{hyperref}
\hypersetup{colorlinks=True,urlcolor=magenta}
\usepackage{url}

\usepackage[accepted]{icml2022}

\usepackage{amsmath}
\usepackage{amssymb}
\usepackage{mathtools}
\usepackage{amsthm}

\usepackage[capitalize,noabbrev]{cleveref}

\theoremstyle{plain}

\theoremstyle{definition}

\theoremstyle{remark}

\usepackage[textsize=tiny]{todonotes}

\icmltitlerunning{What Do We Maximize in Self-Supervised Learning?}

\begin{document}

\twocolumn[
\icmltitle{What Do We Maximize in Self-Supervised Learning?}



\icmlsetsymbol{equal}{*}

\begin{icmlauthorlist}
\icmlauthor{Ravid Shwartz-Ziv}{yyy}
\icmlauthor{Randall Balestriero}{comp}
\icmlauthor{Yann LeCun}{yyy,comp}
\end{icmlauthorlist}

\icmlaffiliation{yyy}{New York Universityy}
\icmlaffiliation{comp}{Meta AI Research}

\icmlcorrespondingauthor{Ravid Shwartz-Ziv}{ravidziv@gmail.com}

\icmlkeywords{Machine Learning, ICML}

\vskip 0.3in
]



\printAffiliationsAndNotice{\icmlEqualContribution} 

\begin{abstract}
In this paper, we examine self-supervised learning methods, particularly VICReg, to provide an information-theoretical understanding of their construction. As a first step, we demonstrate how information-theoretic quantities can be obtained for a deterministic network, offering a possible alternative to prior work that relies on stochastic models. This enables us to demonstrate how VICReg can be (re)discovered from first principles and its assumptions about data distribution. Furthermore, we empirically demonstrate the validity of our assumptions, confirming our novel understanding of VICReg. Finally, we believe that the derivation and insights we obtain can be generalized to many other SSL methods, opening new avenues for theoretical and practical understanding of SSL and transfer learning.
\end{abstract}

\section{Introduction}

Self-Supervised Learning (SSL) algorithms \citep{bromley1993signature} learn representations using a proxy objective (i.e., SSL objective) between inputs and self-defined signals. The results indicate that the learned representations can generalize well to a wide range of downstream tasks \citep{chen2020simple,misra2020self}, even when the SSL objective does not use downstream supervision during training. In SimCLR \cite{chen2020simple}, for example, a contrastive loss is defined between images with different augmentations (i.e., one as input and the other as a self-supervised signal). Then, we take our pre-learned model as a feature extractor and adopt the features to various applications, including image classification, object detection, instance segmentation, and pose estimation \citep{caron2021emerging}. However, despite the success in practice, only a few works \cite{arora2019theoretical, lee2021predicting} provide theoretical insights into the learning efficacy of SSL.

In recent years, information theory methods have played a key role in several notable deep learning achievements, from practical applications in representation learning as the variational information bottleneck \cite{alemi2016deep}, to theoretical investigations (e.g., the generalization bound induced by mutual information \cite{xu2017information, steinke2020reasoning, shwartz2022information}. Moreover, different deep learning problems have been successfully approached by developing and applying novel estimators and learning principles derived from information-theoretic quantities, such as mutual information estimation. Many works have attempted to analyze SSL from an information theory perspective. An example is the use of the mutual information neural estimator (MINE) \cite{belghazi2018mine} in representation learning \cite{hjelm2018learning} in conjunction with the renowned information maximization (InfoMax) principle \cite{linsker1988self}. However, looking at these works may be confusing. Numerous objective functions are presented, some contradicting each other, as well as many implicit assumptions. Moreover, these works rely on a crucial assumption: a stochastic (often Gaussian) DN mapping, which is rarely the case nowadays.

This paper presents a unified framework for SSL methods from an information theory perspective which can be applied to deterministic DN training. We summarize our contributions into two points: (i) Firdt, in order to study deterministic DNs from an information theory perspective, we shift stochasticity to the DN input, which is a much more faithful assumption for current training techniques.
 (ii) Second, based on this formulation, we analyze how current SSL methods that use deterministic networks optimize information-theoretic quantities. 

\section{Background}
\label{sec:background}

{\bf Continuous Piecewise Affine (CPA) Mappings.}~
A rich class of functions emerges from piecewise polynomials: spline operators. In short, given a partition $\Omega$ of a domain $\mathbb{R}^D$, a spline of order $k$ is a mapping defined by a polynomial of order $k$ on each region $\omega \in \Omega$ with continuity constraints on the entire domain for the derivatives of order $0$,\dots,$k-1$. As we will focus on affine splines ($k=1$), we define this case only for concreteness. An $K$-dimensional affine spline $f$ produces its output via
\begin{align}
    f(\bz) = \sum_{\omega \in \Omega}(\bA_{\omega}\bz+\bb_{\omega})\Indic_{\{\bz \in \omega\}},\label{eq:CPA}
\end{align}
with input $\bz\in\mathbb{R}^{D}$ and $\bA_{\omega}\in\mathbb{R}^{K \times D},\bb_{\omega}\in\mathbb{R}^{K},\forall \omega \in \Omega$ the per-region {\em slope} and {\em offset} parameters respectively, with the key constraint that the entire mapping is continuous over the domain $f\in\mathcal{C}^{0}(\mathbb{R}^D)$. Spline operators and especially affine spline operators have been widely used in function approximation theory \cite{cheney2009course}, optimal control \citep{egerstedt2009control}, statistics \citep{fantuzzi2002identification}, and related fields.
\\
{\bf Deep Networks.}~
A deep network (DN) is a (non-linear) operator $f_\Theta$ with parameters $\Theta$ that map a {\em input} $\bx\in\R^D$ to a {\em prediction} $\by\in \R^K$. The precise definitions of DNs operators can be found in \citet{goodfellow2016deep}. We will omit the $\Theta$ notation for clarity unless needed. The only assumption we require for our study is that the non-linearities present in the DN are CPA, as is the case with (leaky-) ReLU, absolute value, and max-pooling. In that case, the entire input-output mapping becomes a CPA spline with an implicit partition $\Omega$, the function of the weights and architecture of the network \citep{montufar2014number,balestriero2018spline}. For smooth nonlinearities, our results hold from a first-order Taylor approximation argument. 
\\
{\bf  Self-Supervised Learning}. Joint embedding methods learn the DN parameters $\Theta$ without supervision and input reconstruction. 
Due to this formulation, the difficulty of SSL is to produce a good representation for downstream tasks whose labels are not available during training ---while avoiding a trivially simple solution where the model maps all inputs to constant output. Many methods have been proposed to solve this problem. {\em Contrastive methods} learn representations by contrasting positive and negative examples, e.g. SimCLR \cite{chen2020simple} and its InfoNCE criterion \cite{oord2018representation}.
Other recent work introduced {\em non-contrastive methods} that employ different regularization methods to prevent collapsing of the representation. Several papers used stop-gradients and extra predictors to avoid collapse \cite{chen2021exploring, grill2020bootstrap} while \citet{caron2020unsupervised} uses an additional clustering step. As opposed to contrastive methods, noncontrastive methods do not explicitly rely on negative samples. Of particular interest to us is the {\em VICReg} method \cite{bardes2021vicreg} that considers two embedding batches $\bZ=\left[f(\bx_1),\dots,f(\bx_N)\right]$ and $\bZ^\prime=\left[f(\bx'_1),\dots,f(\bx'_N) \right]$ each of size $(N \times K)$. Denoting by $\bC$ the $(K \times K)$ covariance matrix obtained from $[\bZ,\bZ']$ we obtain the VICReg triplet loss
\begin{gather*}
    \mathcal{L}\hspace{-0.1cm}=\frac{1}{K}\sum_{k=1}^K\hspace{-0.1cm}\left(\hspace{-0.1cm}\alpha\max \left(0, \gamma- \sqrt{\bC_{k,k} +\epsilon}\right)\hspace{-0.1cm}+\hspace{-0.1cm}\beta \sum_{k'\neq k}\hspace{-0.1cm}\left(\bC_{k,k'}\right)^2\hspace{-0.1cm}\right)\\
    \;\;\;+\gamma\| \bZ-\bZ'\|_F^2/N.
\end{gather*}
Our goal will now be to formulate SSL as an information-theoretic problem from which we can precisely relate VICReg to known methods even with a deterministic network.

{\bf Deep Networks and Information-Theory.}~
Recently, information-theoretic methods have played a key role in several remarkable deep learning achievements \citep{alemi2016deep, xu2017information, steinke2020reasoning, shwartz2017opening}. Moreover, different deep learning problems have been successfully approached by developing and applying information-theoretic estimators and learning principles \cite{hjelm2018learning, belghazi2018mine, piran2020dual, shwartz2018representation}. There is, however, a major problem when it comes to analyzing information-theoretic objectives in deterministic deep neural networks: the source of randomness. The mutual information between the input and the representation in such networks is infinite, resulting in ill-posed optimization problems or piecewise constant, making gradient-based optimization methods ineffective \citep{amjad2019learning}. To solve these problems, researchers have proposed several solutions. For SSL, stochastic deep networks with variational bounds could be used, where the output of the deterministic network is used as parameters of the conditional distribution \citep{lee2021compressive,shwartz2020information}. \citet{dubois2021lossy} suggested another option, which assumed that the randomness of data augmentation among the two views is the source of stochasticity in the network. For supervised learning, \citet{2018Estimating} introduced an auxiliary (noisy) DN framework by injecting additive noise into the model and demonstrated that it is a good proxy for the original (deterministic) DN in terms of both performance and representation. Finally, Achille and Soatto \cite{achille2018information} found that minimizing a stochastic network with a regularizer is equivalent to minimizing cross-entropy over deterministic DNs with multiplicative noise. However, all of these methods assume that the noise comes from the model itself, which contradicts current training methods. In this work, we explicitly assume that the stochasticity comes from the data, which is a less restrictive assumption and does not require changing current algorithms.

\section{Information Maximization of Deep Networks Outputs}
\label{sec:theory}

This section first sets up notation and assumption on the information-theoretic challenges in self-supervised learning (\cref{sec:info_max}) and on our assumptions regarding the data distribution (\cref{sec:manifold}) so that any training sample $\bx$ can be seen as coming from a single Gaussian distribution as in $\bx \sim \mathcal{N}(\mu_{\bx},\Sigma_{\bx})$. From this we obtain that the output of any deep network $f(\bx)$ corresponds to a mixture of truncated Gaussian (\cref{sec:output_density}). In particular, it can fall back to a single Gaussian under some small noise ($\det(\Sigma) \rightarrow \epsilon$) assumptions. These results will enable information measures to be applied to deterministic DNs. We then recover the known SSL methods \cite{bardes2021vicreg} by making different assumptions about the data distribution and estimating their information.

\subsection{SSL as an Information-Theoretic Problem}
\label{sec:info_max}

To better grasp the difference between key SSL methods, we first formulate the general SSL goal from an information-theoretical perspective.

We start with the\textit{ MultiView InfoMax} principle,  i.e., maximizing the mutual information between the representations of the two views. To do so, as shown in \citet{federici2020learning}, we need to maximize $I(Z;X^\prime)$ and $I(Z^\prime;X)$. We can do so by a lower bound using
\begin{align}
    \label{eq:lower_bound3}
    I(Z,X^\prime) \hspace{-0.1cm}=\hspace{-0.1cm} H(Z) \hspace{-0.1cm}-\hspace{-0.1cm} H(Z|X^\prime) \hspace{-0.1cm}\geq \hspace{-0.1cm}H(Z) \hspace{-0.1cm}+\hspace{-0.1cm}E[\log q(z|x^\prime)] 
\end{align}
where $H(Z)$ is the entropy of $Z$. In supervised learning, where we need to maximize $I(Z;Y)$, the labels ($Y$) are fixed, the entropy term $H(Y)$ is constant, and you only need to optimize the log-loss $E[\log q(z|x)]$ (cross-entropy or square loss). However, in SSL, the entropy $H(Z)$ and $H(Z^\prime)$ are not constant and can be optimized throughout the learning process. Therefore, only maximizing $E[\log q(z|x^\prime)]$ will cause it to collapse to the trivial solution of making the representations constant (where the entropy goes to zero). To regularize these entropies, i.e., prevent collapse, different methods utilize different approaches to implicit regularizing information. To recover them in \cref{sec:recovery}, we must first introduce the notation and results around the data distribution (\cref{sec:manifold}) and how a DN transforms that distribution (\cref{sec:output_density}).


\subsection{Data Distribution Hypothesis}

\label{sec:manifold}
Our first step is to assess how the output random variables of the network are represented, assuming a distribution on the data itself.  
Under the manifold hypothesis, any point can be seen as a Gaussian random variable with a low-rank covariance matrix in the direction of the manifold tangent space of the data. Therefore, we will consider throughout this study the conditioning of a latent representation with respect to the mean of the observation, i.e., $X|\bx^* \sim \mathcal{N}(\bx^*,\Sigma_{\bx^*})$ where the eigenvectors of $\Sigma_{\bx^*}$ are in the same linear subspace than the tangent space of the data manifold at $\bx^*$, which varies with the position of $\bx^*$ in space.

Hence a dataset is considered to be a collection of $\{\bx^*_n,n=1,\dots,N\}$ and the full data distribution to be a sum of low-rank covariance Gaussian densities as in
\begin{equation}
X \sim \sum_{n=1}^N\mathcal{N}(\bx^*_n,\Sigma_{\bx^*_n})^{T=n},T\sim {\rm Cat}(N),
\end{equation}
with $T$ the uniform Categorical random variable. To keep things simple and without loss of generality, we consider that the effective support of $\mathcal{N}(\bx^*_i,\Sigma_{\bx^*_i})$ and $\mathcal{N}(\bx^*_j,\Sigma_{\bx^*_j})$ do not overlap. This keeps things general, as it is enough to cover the domain of the data manifold overall, without overlap between different Gaussians. 
Hence, in general, we have that.
\begin{equation}
p(\bx) \approx \mathcal{N}\left(\bx;\bx^*_{n(\bx)},\Sigma_{\bx^*_{n(\bx)}}\right)/N,\label{eq:x_density}
\end{equation}
where $\mathcal{N}\left(\bx;.,.\right)$ is the Gaussian density at $\bx$ and
with $n(\bx)=\argmin_n (\bx-\bx^*_n )^T\Sigma_{\bx^*_n}(\bx-\bx^*_n )$.
This assumption that a dataset is a mixture of Gaussians with nonoverlapping support will simplify our derivations below, which could be extended to the general case if needed. Note that this is not restrictive since, given a sufficiently large $N$, the above can represent any manifold with an arbitrarily good approximation.

\subsection{Data Distribution After Deep Network Transformation}
\label{sec:output_density}

Consider an affine spline operator $f$ (Eq.~\ref{eq:CPA}) that goes from a space of dimension $D$ to a space of dimension $K$ with $K\geq D$. 
The span, that we denote as image, of this mapping is given by
\begin{equation}
    Im(f)\triangleq\{f(\bx):\bx \in \mathbb{R}^D\}=\bigcup_{\omega \in \Omega} \Aff(\omega;\bA_{\omega},\bb_{\omega})\label{eq:generator_mapping}
\end{equation}
with $\Aff(\omega;\bA_{\omega},\bb_{\omega})=\{\bA_{\omega}\bx+\bb_{\omega}:\bx\in \omega\}$ the affine transformation of region $\omega$ by the per-region parameters $\bA_{\omega},\bb_{\omega}$, and with $\Omega$ the partition of the input space in which $\bx$ lives in.
We also provide an analytical form of the per-region affine mappings in \cref{sec:background}. Hence, the DN mapping consists of affine transformations on each input space partition region $\omega \in \Omega$ based on the coordinate change induced by $\bA_{\omega}$ and the shift induced by $\bb_{\omega}$.

When the input space is equipped with a density distribution, this density is transformed by the mapping $f$. In general, finding the density of $f(X)$ is an intractable task. However, given our disjoint support assumption provided in \cref{sec:manifold}, we can arbitrarily increase the representation power of the density by increasing the number of prototypes $N$. In doing so, the support of each Gaussian is included with the region $\omega$ in which its means lie in, leading to the following result.

\begin{thm}
\label{cor:mixture}
Given the setting of \cref{eq:x_density} the unconditional DN output density denoted as Z is a mixture of the affinely transformed distributions $\bx|\bx^*_{n(\bx)}$ e.g. for the Gaussian case
$$Z\hspace{-0.1cm} \sim\hspace{-0.1cm} \sum_{n=1}^{N}\mathcal{N}\hspace{-0.1cm}\left(\hspace{-0.1cm}\bA_{\omega(\bx^*_{n})}\bx^*_{n}+\bb_{\omega(\bx^*_{n})},\bA^T_{\omega(\bx^*_{n})}\Sigma_{\bx^*_{n}}\bA_{\omega(\bx^*_{n})}\hspace{-0.1cm}\right)^{T=n},$$
where $\omega(\bx^*_{n})=\omega \in \Omega \iff \bx^*_{n} \in \omega$ is the partition region in which the prototype $\bx^*_{n}$ lives in.
\end{thm}

The proof of the above involves the fact that if $\int_{\omega}p(\bx|\bx^*_{n(\bx)})d\bx \approx 1$ then $f$ is linear within the effective support of $p$. Therefore, any sample from $p$ will almost surely lie within a single region $\omega \in \Omega$ and therefore the entire mapping can be considered linear with respect to $p$. Thus, the output distribution is a linear transformation of the input distribution based on the per-region affine mapping.

\section{Information Optimization and Optimality}
\label{sec:recovery}

Based on our analysis, we now show how specific SSL algorithms can be derived. According to Section \ref{sec:info_max}, we want to maximize $I(Z;X^\prime)$ and $I(Z^\prime;X)$.  When our input noise is small, we can reduce the conditional output density $p_{\bz|\bx*}$ to a single Gaussian.

$$(Z^\prime|X^\prime=x_n) \sim \mathcal{N} \left(\mu_n, \Sigma_{n} \right),$$
where we abbreviated the parameters.
Using that, and the result from \cref{sec:info_max},  we see that one should optimize both $H(Z|X^\prime)$ and $H(Z)$. As in a standard regression task, we assume a Gaussian observation model, which means $p(z|z^\prime) \sim \mathcal{N}(z^\prime, \Sigma_r)$. Using the mean square error as a loss function in regression tasks is a particular application of this assumption, where $\Sigma_r=I$. To compute the expected loss, we need to marginalize out the stochasticity in $Z^\prime$, which means that the conditional decoding map is a Gaussian:
\begin{equation}
    q(z|X^\prime=x^*_{n}) \hspace{-0.1cm}=\hspace{-0.1cm} \int  q(z|z^\prime)p(z^\prime|x^*_{n}) dz^\prime,
\end{equation}
which gives the distribution $\mathcal{N}(\mu_{n}, \Sigma_r + \Sigma_{n})$
meaning that we can lower bound the mutual information with
\begin{align}
\label {eq:izz_bound}
  I(&Z;X^\prime)  \geq H(Z) + E[\log(q(z|x^\prime)] \hspace{-0.1cm}=\hspace{-0.1cm}  H(Z) -\frac{d}{2}\log 2\pi \Sigma_r  \nonumber\\ 
  &-\sum_{n=1}^N \frac12 (z_n-z_n^\prime)^T\Sigma_{r}^{-1}(z_n-z_n^\prime) -\log |\Sigma{_n}|.
\end{align}
What happens if we attempt to optimize this objective? The only intractable component is the entropy of $Z$. We will begin by examining $Z$ itself. It is natural to ask why the entropy of $Z$ will not increase to infinity. Intuitively, the answer is that $H(Z)$ and $H(Z|X^\prime)$ are tied together, and one cannot increase without the other. Now, recalling that under our distribution assumption, $Z$ is a mixture of Gaussian (recall Thm.~\ref{cor:mixture}), we can see how the existing upper and lower bounds could be used for this case; for example, the ones in \citet{boundsgmmentropy}.

\begin{figure}[t!]
    \centering
    \includegraphics[width=\linewidth]{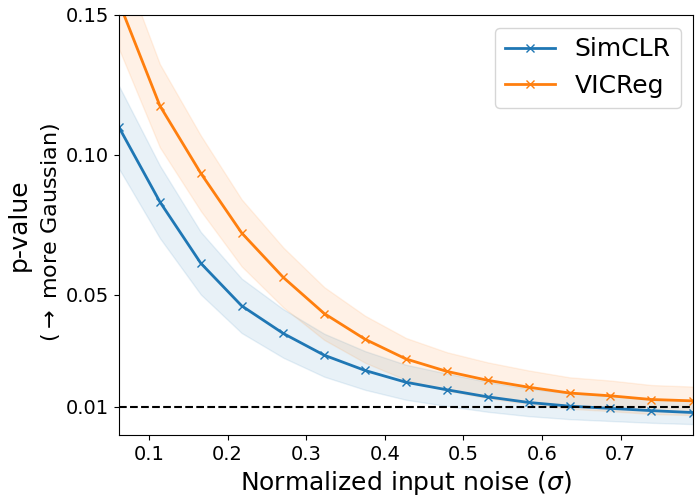}\\
    \caption{\textbf{The network output with VICReg training is more gaussian for small input noise}. The P-value of the normality test for different SSL models trained on CIFAR-10 for different input noise levels. The x-axis is the coefficient that multiplies the data distribution standard deviation to obtain the Gaussian standard deviation that samples around each image. The dashed line represents the point at which the null hypothesis (Gaussian distribution of the network output) can be rejected with $99\%$ confidence.}
    \label{fig:gaussian}
\end{figure}


\subsection{Deriving VICReg From First Principles}
\label{sec:vicreg}

We now propose to recover VICReg from the first principles per the above information-theoretic principle.  

Recall that our goal is to estimate the entropy $H(Z)$ in \cref{eq:izz_bound}, where $Z$ is a Gaussian mixture. This quantity is not known for a mixture of Gaussians due to the logarithm of a sum of exponential functions, except for the special case of a single Gaussian density. There are, however, several approximations in the literature that include both upper and lower bounds. Among the methods, some use the logarithmic sum of the probability \cite{kolchinsky2017estimating}, and some use entropy-adjusted logarithmic probabilities \cite{entropyapprox2008}.

An even simpler solution is to approximate the entire mixture as a single Gaussian by capturing only the first two moments of the mixture distribution. Since the Gaussian distribution maximizes the entropy for a given covariance matrix,  this method provides an upper bound approximation of our entropy of interest $H(Z)$. In this case, denoting by $\Sigma_{Z}$ is the covariance matrix of $Z$, we find that we should maximize the following objective:

\begin{align*}
  \max_{Z,Z^\prime} & \hspace{-0.1cm}\sum_{n=1}^N\log \frac{|\Sigma_{Z}|}{ |\Sigma_{n}|} \hspace{-0.1cm}-\hspace{-0.1cm} \frac12  (z_n-z_n^\prime)^T\Sigma_{r}^{-1}\hspace{-0.03cm}(z_n-z_n^\prime) \hspace{-0.1cm} -\frac{d}{2}\log 2\pi \Sigma_r
\end{align*}
where $\Sigma_{r}$  is constant with respect to our optimization process, and the second term is the prediction performance of one representation from the other. 
A key result from \citet{shi2009data} connects the eigenvectors and eigenvalues of $\Sigma_{Z}$ and those of each component $\Sigma_{i},\forall i$, and showed that under the assumption that the separation $(\mu_i - \mu_j)\Sigma_i^{-1}(\mu_i - \mu_j)^T$ between the different components is large enough ---which holds true in our case as per our data distribution model--- the eigenfunctions of $\Sigma_{i}, \forall i$ are approximately the eigenfunctions of $\Sigma_{Z}$. Therefore, in our case, 
this means that since all those eigenvalues are tied,
we only need to find the most efficient way to maximize $|\Sigma_{Z}|$.

We know that the determinant of the matrix is the product of its eigenvalues. For every positive matrix, the maximum eigenvalue is greater than or equal to each diagonal element. Therefore, under the constraint of the eigenvalues of the matrix, the most efficient way is to decrease the off-diagonal terms and increase the diagonal terms. By setting $\Sigma_r=I$, {\em we therefore fully recover the VICReg objective}. 

\subsection{SimCLR vs VICReg}
\citet{lee2021compressive} connected the SimCLR objective \cite{chen2020simple} to the variational bound on the information between representations \cite{poole2019variational}, by using the von Mises-Fisher distribution as the conditional variational family. Based on our analysis in Section \ref{sec:vicreg}, we can identify two main differences between SimCLR and VICReg: (i) \textbf{The  conditional distribution $\bp(\bz|\bx^\prime$)}; SimCLR assumes a von Mises-Fisher distribution for the encoder, while VICReg assumes a Gaussian distribution. (ii) \textbf{Entropy estimation}; the entropy term in SimCLR, $H(Z)=\int p(z|x^\prime)p(x^\prime)dx^\prime$, is approximate based on the finite sum of the input samples. VICReg, however, uses a different approach and estimates the entropy of $Z$ only from the first second moments. Creating self-supervised methods that combine these two differences would be an interesting future research direction.  In theory, none of these assumptions is more valid than the other, and it depends on the specific task and our computational constraints.

\subsection{Empirical Evaluation }

The next step is to verify the validity of our assumptions. Based on the theory presented in Section \ref{sec:output_density}, the conditional output density $p_{\bz|x=i}$ reduces to a single Gaussian with decreasing input noise. We validated it using a ResNet-18 model trained with SimCLR or VICReg on the CIFAR-10 dataset \cite{Krizhevsky09learningmultiple}. From the test dataset, we sample $512$ Gaussian samples for each image and analyzed whether these samples remain Gaussian (for each image) at the penultimate layer of the DN, that is, before the linear classification layer, independently for each output dimension. Then, we employ D'Agostino and Pearson's test \cite{10.2307/2334522} to compute the p-value of the normality test under the null hypothesis that the sample represents a normal distribution. In this test, the deviation from normality is measured, and the test aims to determine whether the sample represents a normally distributed population. A kurtosis and skewness transformation is used to perform the test. The process is repeated for different noise standard deviations. Figure \ref{fig:gaussian} shows the p-value as a function of the normalized standard deviation. We can observe that the network's output is indeed Gaussian with a high probability for small input noise. As we increase the input noise, the network's output becomes less Gaussian until the noise distribution can be rejected with $99\%$ confidence. Moreover, we can see that VICReg is, interestingly, more "Gaussian" than SimCLR, which may have to do with the fact that it optimizes only the second moments of the density distribution to regularize $H(Z)$.


\section{Conclusions}
\label{sec:conclusion}

In this study, we examine SSL's objective function from an information-theoretic perspective. Our analysis, based on transferring the required stochasticity to the input distribution, shows how to derive SSL objectives. Therefore, it is possible to obtain an information-theoretic analysis even when using deterministic DNs. In the second part, we rediscovered VICReg's loss function from first principles and showed its implicit assumptions. In short, VICReg performs a crude lower bound estimate of the output density entropy by approximating this distribution with a Gaussian matching the first two moments. Finally, we empirically validated that our assumptions are valid in practice, thus confirming the validity of our novel understanding of VICReg.   Our work opens many new paths for future research; A better estimation of information-theoretic quantities fits our assumptions. Another exciting research direction is to identify which SSL method is preferred based on data properties.

\bibliography{references}
\bibliographystyle{icml2022}

\appendix

\end{document}